\documentclass{egpubl}
\usepackage{ceig2017}
\usepackage{amsmath}
\usepackage{algorithm}
\usepackage{algpseudocode}
\usepackage{tabularx} 
\usepackage[normalem]{ulem}
\usepackage[table,xcdraw]{xcolor}
\usepackage[normalem]{ulem}  

\newcommand{\IGNORE}[1]{}

\newcommand{\noopsort}[1]{}     

\newdimen\boxtextwidth
 \WsPaper         
 \electronicVersion 

\ifpdf \usepackage[pdftex]{graphicx} \pdfcompresslevel=9
\else \usepackage[dvips]{graphicx} \fi

\PrintedOrElectronic

\usepackage{t1enc,dfadobe}

\usepackage{egweblnk}
\usepackage{cite}

\title%
{Transfer Learning for Illustration Classification}

\author[Manuel Lagunas \& Elena Garces]
{\parbox{\textwidth}
	{\centering 
		Manuel Lagunas$^{1}$
	    Elena Garces$^{2}$
	}
	\\
	{\parbox{\textwidth}{\centering 
			$^1$Universidad de Zaragoza, I3A \\
			$^2$Technicolor
		}
	}
}

\begin{document}
	 \teaser{
	  \includegraphics[width=0.95\linewidth]{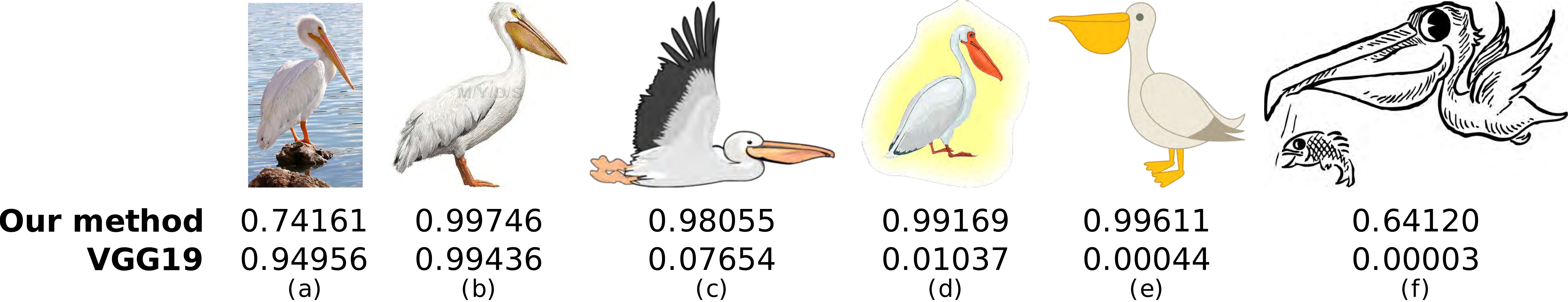}
\centering
\caption{Comparison of the probabilities of the images that belong to the class \textbf{pelican} using our method and the network VGG19~\cite{Simonyan14}. 
We can see how the network VGG19 pre-trained on ImageNet~\cite{Russakovsky15} is capable of classifying correctly the images (a)-(b). Note that image (a) is a photograph and image (b) is an illustration which has similar colours, gradients and edges than the natural image. On the contrary, VGG19 fails on images (c)-(f), with a more cartoon-like style. Our method is able to predict correctly the class of all images independently of the depicted style. We can observe that the certainty of the predictions decreases as the degree of abstraction of the images increases.
}
	 \label{fig:teaser}
	}
	
	\maketitle

	\def\UrlBreaks{\do\/\do-}

	\begin{abstract}
		
		The field of image classification has shown an outstanding success thanks to the development of deep learning techniques. Despite the great performance obtained, most of the work has focused on natural images ignoring other domains like artistic depictions. In this paper, we use transfer learning techniques to propose a new classification network with better performance in illustration images. Starting from the deep convolutional network VGG19, pre-trained with natural images, we propose two novel models which learn object representations in the new domain. 
Our optimized network will learn new low-level features of the images (colours, edges, textures) while keeping the knowledge of the objects and shapes that it already learned from the ImageNet dataset. Thus, requiring much less data for the training.
		We propose a novel dataset of illustration images labelled by content where our optimized architecture achieves  \textbf{86.61\%} of top-1 and \textbf{97.21\%} of top-5 precision. We additionally demonstrate that our model is still able to recognize objects in photographs. 		
%

	\end{abstract}
	
	\section{Introduction}
\label{sc:intro}

The ability of the human being to identify and recognize objects and textures is unquestionable. In practice, humans are able to recognize almost any object in a photograph or a picture regardless of the illumination, the perspective, the style, or even the level of abstraction in a drawing.  However, computers are not as developed and, only until recently, the precision rates of classifying objects in natural images were not even close to the human level.
The emergence of deep learning techniques in 2012 was a major revolution in the computer vision field, in particular, for image classification, reaching accuracy rates of more than 95\%. These techniques, although very compelling for natural images, barely explore another dimension of our perception which is the pictorial level. 

One of the keys to the success of these networks was the availability of hundreds of thousands of annotated natural images and curated datasets~\cite{Russakovsky15, Krizhevsky-cifar, Everingham15}, which allowed to learn very complex and non-linear pixel statistics, relationships and patterns. However, training these networks is a very expensive task in terms of time and resources. Thus, training a deep neural network from scratch requires very large amounts of annotated data and great computational power. To overcome this problem, \emph{transfer learning} techniques aim to use existing pre-trained architectures and make them useful for a new dataset by retraining them with much less data or classifying their high-level layers with simpler machine learning algorithms.

In this work, we want to explore the pictorial domain, particularly in illustration pictures, for the task of image classification. We rely on the intuition that at a local level, illustration depictions have statistics in strokes, edges, or textures, very different to those found on natural images. 
However, at a higher level, the essential parts that make up the objects like their shapes remain closely the same.

We start using the publicly available deep neural network VGG-19~\cite{Simonyan14}
that was trained on the natural image dataset ImageNet~\cite{Russakovsky15} containing over 1.2 million images. 
First, we evaluate such network with our novel dataset of illustration images labelled by content. Noticing a poor performance, we propose an adaptive layer-based optimization strategy that modifies only a few layers of the network to let it capture better the new content. 
Thus, we propose to restart and train the layers that capture the low-level characteristics of the images~\cite{Zeiler10, Zeiler14}, since those are making the difference with respect to natural images while keeping similar the higher-level layers. 

The contributions of this paper are the following:

\begin{itemize}
	\item[--] We present a new dataset of illustration images labelled by content. 
	\item[--] We evaluate the performance of existing architectures~\cite{Simonyan14} with our new dataset.   
	\item[--] We propose two novel models based on transfer learning techniques~\cite{Oquab14, Lin11, Babenko14, Razavian14} optimized for our data. The first model leverages traditional machine learning techniques and requires a small amount of new data for the training. The second and optimized model requires a larger dataset for the training, but leverages the information already available in the original network, thus requiring much less amount of data that would have been necessary if we trained the network from scratch.     
	\item[--] We demonstrate on a small set of natural images that despite the changes in the architecture, the new network is still able to classify natural images accurately.
	
\end{itemize}

	\section{Related work}
\label{sc:related}



The growth of the curated data available together with the rise of the computational power have allowed to develop complex models that outperform humans in a variety of tasks~\cite{He15}. In particular, convolutional neural networks (CNN)~\cite{Lecun89, Lecun01} have shown great results in a task such as image classification. Since Krizhevsky et al.~\cite{Krizhevsky12} presented their deep neural network to the ImageNet~\cite{Russakovsky15} competition, CNNs have become the main resource for solving image classification tasks~\cite{Simonyan14, He15, Szegedy15}.

Training a deep neural network from scratch takes a lot of time and resources. Therefore, it is common to take an existing pre-trained model~\cite{Everingham15, Krizhevsky-cifar, Russakovsky15} and adapt it to a new dataset. This technique is called \emph{transfer learning}. People have studied the problem of transfer learning for different tasks. Within the context of natural image classification, it has been shown that restarting the fully-connected layers of a convolutional neural network, such as AlexNet~\cite{Krizhevsky12}, and training them with the new dataset yield successful results~\cite{Oquab14}. Other approaches suggest that top layers of large convolutional neural networks are powerful high-level image descriptors (called \textit{neural codes})~\cite{Babenko14}. Further work~\cite{Sermanet13} explore how the \textit{neural codes} obtained from pretrained neural networks can be used in conjunction with support vector machines and dimensionality reduction techniques to achieve accurate results in new datasets of the same image domain~\cite{Razavian14, Lin11, Donahue13, Tang13}.

Yosinki et al.~\cite{Yosinski14}, explore how transferable are features in deep neural networks. They make an experimental setup to benchmark a group of transfer learning techniques against different datasets. They prove that the effectiveness of these techniques decreases as the initial and target datasets become less similar.

Similar to ours is the work of Crowley and Zisserman~\cite{Crowley14a}, which explore the transfer learning problem with images of a different domain. They utilize a dataset of paintings and a pretrained network. They extract the image descriptors and use a support vector machine as the classifier obtaining great results. Although they are using paintings as the input, its low-level features are not remarkable because their style is realistic and resemble natural images. The dataset complexity lies in the differences of the paintings over time. There has also been interest in transfer learning between different image domains for medical image classification~\cite{Cheplygina17, Christodoulidis16}.
Our dataset is made of illustration images that have more degrees of abstraction than realistic paintings. This together with the differences in the low-level features, complicates the detection of relevant characteristics of the image by the network.

	\section{Overview}
\label{sc:overview}

Our goal is to find a model that is able to correctly predict class labels for illustration and clip art data. There are a number of curated datasets which contain labelled images of real objects, like the ImageNet dataset~\cite{Russakovsky15} which contains more than 1.2 million pictures. However, there is not a proper dataset for this kind of cartoon-like styles that we aim to analyze. Hence, we first created a dataset of illustration images labelled by content (Section~\ref{sc:dataset}). This dataset is composed of two sets of data that will be used for different tasks. The \textit{noisy} dataset, with more than 180K images separated in 826 classes and the \textit{curated} dataset, with more than 4k images and 23 classes. Both, the curated and noisy dataset are split into a fixed sets of 
training, evaluation and testing data.

\begin{figure}[h!]
	\centering
	\includegraphics[width=0.85\linewidth]{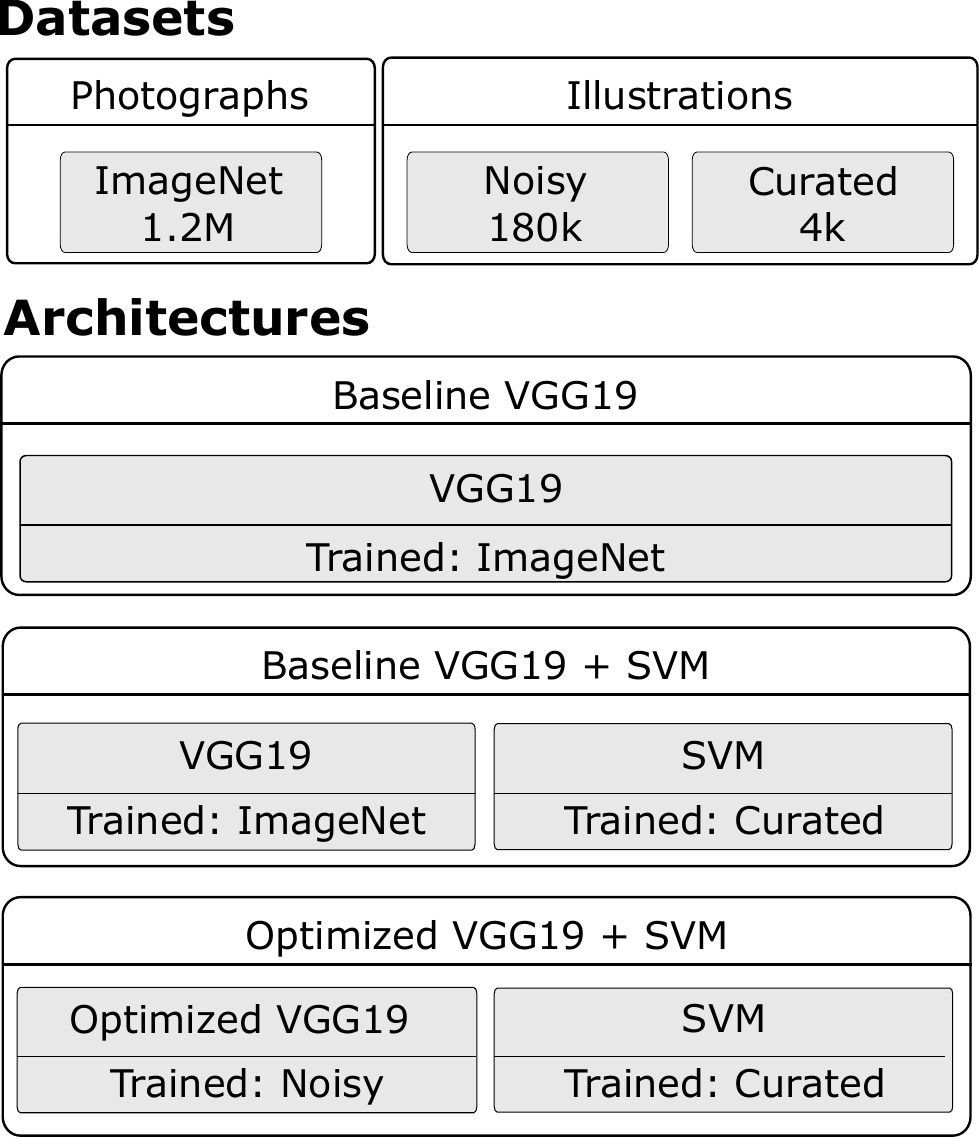}
	\caption{ 
		Overview of our work. It contains two main blocks the \textbf{datasets} and the \textbf{architectures}. The datasets used have two types of images: natural images (ImageNet) and illustrations (Noisy and Curated) and were used to train the proposed architectures. The main architectures are: the network VGG19~\cite{Simonyan14} pre-trained on ImageNet (Baseline VGG19), the VGG19 combined with a SVM trained on the Curated dataset (Baseline VGG19 + SVM) and the VGG19 optimized with the Noisy dataset combined with a SVM trained on the Curated dataset (Optimized VGG19 + SVM).}
	\label{fig:overview}
\end{figure}

We first evaluate the existing VGG19~\cite{Simonyan14} deep neural network, which has proven to perform very well predicting classes in natural images. In Section~\ref{sc:vgg19} we provide a summary of this 
architecture and shows its performance in our data. 
Since the accuracy obtained is quite low, we consider it as the baseline (\textit{Baseline VGG19}) and propose two novel models inspired by transfer learning techniques~\cite{Razavian14, Lin11, Oquab14}.  
In the first model (\textit{Baseline VGG19 + SVM}), explained in Section~\ref{sc:method-1}, we use a SVM to classify features extracted from the deep network VGG19. 
The performance increases with respect to the previous architecture but remains low. Hence, we propose a second model (\textit{Optimized VGG19 + SVM}), described in Section~\ref{sc:method-2}, which is based on two steps: first, we perform an adaptive layer-based optimization using our \textit{noisy} dataset; then, as before, we extract the features of the optimized network and train a SVM using our \textit{curated} dataset.
This model produces accuracy rates of 86.61\% in precision top-1 and 97.21\% in top-5. Improving the previous architecture by a 20\% and 10\% in precision top-1 and top-5 respectively.
A summary of the components of our work is shown in Figure~\ref{fig:overview}.

	\section{Gathering data}
\label{sc:dataset}

Large amount of images labelled accurately can be easily accessed thanks to datasets like ImageNet~\cite{Russakovsky15} or CIFAR~\cite{Krizhevsky-cifar} among other. Unfortunately, most of the curated data available online mostly contain natural images, making the process more complicated when working with other image domains. 

\subsection{Mapping images to ImageNet classes}

\begin{figure}[htb]
	\includegraphics[width=\linewidth]{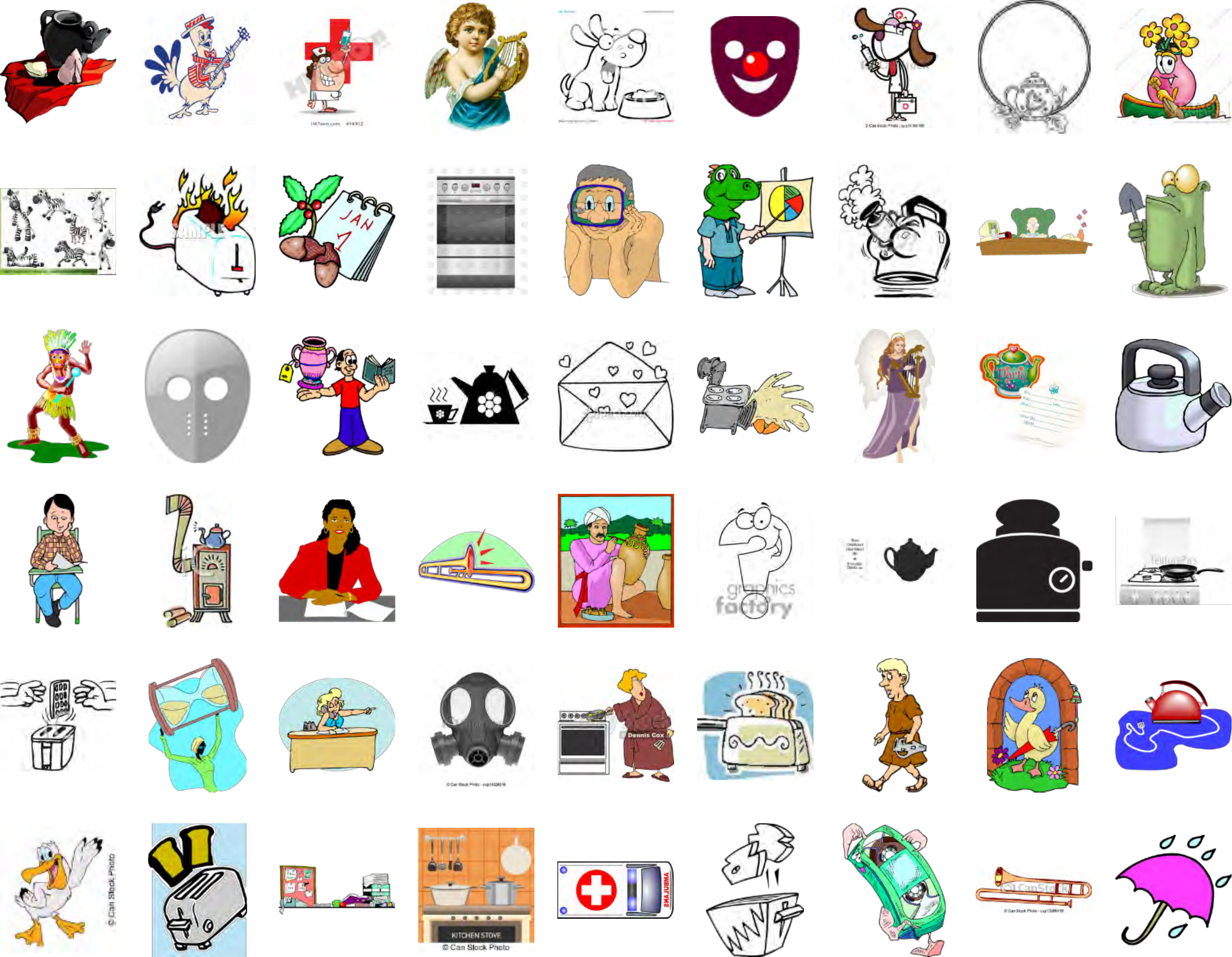}
	\caption{Randomly sampled images from different classes belonging to the illustration dataset used in this work.}
	\label{fig:clipartDataset}
\end{figure}

In this paper, we use an illustration dataset obtained from the work of Garces et al.~\cite{Garces14}. This dataset contains more than 200k clip art images of different objects (see Figure~\ref{fig:clipartDataset}). 
Each image is tagged with a keyword which identifies the object and/or the action e.g. \textit{dog-running, ambulance-1, dog-eating}, etc. In order to compare our performance with existing networks, we need to find the mapping between the image names and the ImageNet existing classes.
ImageNet is a database that holds more than 1.2 million of images within 1000 classes, every year they do a competition to test the performance of the presented models with this data.
The algorithm to map the illustrations to the ImageNet classes consists of a search throughout the class and image names where both are tokenized and their stop-words removed. After that, each word inside the image name is compared with each word in the class name (it could be that images or class names are compound words), if one of the words inside the name or the class matches, the image is copied to the corresponding class.


\subsection{Curating the dataset}

After mapping the image names to ImageNet classes, we obtain a new dataset with around 180K images and 826 classes that we will call \textbf{noisy}. Due to the weak restriction used to create it, we can find problems in its labels:
\begin{itemize}
	\item[--] Ambiguity problems in compound class or image names e.g. a image with name \textit{dog-1.png} could be copied in the class \textit{hot-dog}. 
	\item [--]Problems with class names that have synonyms, e.g. the class \textit{crane} contains images of birds and construction machines. 
\end{itemize}

Using the network VGG19~\cite{Simonyan14} we classify the clip arts and obtain the top-1 and top-5 precision of each class. We use these measurements to select the classes with the highest precision that do not have problems concerning meaning or ambiguity; selecting 23 classes. Some of the 23 classes obtained do not have enough number of samples. In order to increase the number of images in each of the selected class, we use web-scraper which queries Google Images with the class name and download the first 200 clip art images returned.
The web-scrapping process could have copied wrong images to each class, that's why we manually curate them, removing the entries that do not correspond to the name of the class. We obtain a new dataset that is named \textbf{curated} with more than 4000 images and 23 classes. The class names, together with the number of images per class are shown in Table~\ref{tab:curated}.

At the end of the process we obtain two datasets that are mapped to the class names found in ImageNet~\cite{Russakovsky15}:
\begin{itemize}
	\item \textbf{Noisy}: It contains the images directly extracted from the dataset used by Garces et al~\cite{Garces14} and mapped to ImageNet class names. Some of their images are incorrectly labelled. It contains more than 180k images in 826 classes. 
	\item \textbf{Curated}: It has the classes of the noisy dataset that do not present any problems and that scored the best precision when classified with the network VGG19~\cite{Simonyan14}. Its number of entries has been increased using web-scrapping and each class has been manually curated. At the end, the dataset has 4096 images distributed in 23 classes.
\end{itemize}

All the datasets are split in order to train and validate with 80\% of the data and with 20\% for testing purposes.
\begin{table}[!ht]
\centering
\begin{tabular}{|l|*{4}{c}|}
	\hline
	Class & Ambulance & Banjo & Cassete & Desk \\
	\hline
	Images & $ 144 $ & $ 122 $ & $ 134 $ & $ 458 $ \\
	\hline\hline
	Class & Envelope & Goblet & Hammer & Harp\\
	\hline
	Images & $ 132 $ & $ 135 $ & $ 176 $ & $ 127 $ \\
	\hline\hline
	Class & Hourglass & Jellyfish & Mask & Mosque \\
	\hline
	Images & $ 135 $ & $ 115 $ & $ 278 $ & $ 120 $ \\
	\hline\hline
	Class & Pelican & Printer & Shovel & Stove\\
	\hline
	Images & $ 150 $ & $ 235 $ & $ 155 $ & $ 174 $ \\
	\hline\hline
	Class  & Syringe & Teapot  & Toaster & Trombone \\
	\hline
	Images  & $ 152 $ & $ 199 $& $ 185 $ & $ 141 $\\
	\hline\hline
	Class  & Umbrella & Vase & Zebra & \multicolumn{1}{|c|}{\textbf{global}}\\
	\hline
	Images  & $ 254 $ & $ 244 $ & $ 126 $ & \multicolumn{1}{|c|}{\textbf{4091}} \\
	\hline
\end{tabular}
\caption{Number of images in each class of the Curated dataset. The number of images has been obtained after mapping each clipart image to a ImageNet class, selecting the best classes by classifying them with the VGG19 network, increasing the number of images using a web-scrapping technique on Google Images and manually curating each class.}
\label{tab:curated}
\end{table}

	\section{Baseline VGG19}
\label{sc:vgg19}

In this section, we present the basic concepts of deep convolutional neural networks (CNN). First, we briefly describe the architecture used as the baseline, VGG19~\cite{Simonyan14}. Then, we evaluate its performance with our datasets where we show how the accuracy of the network drops when the target and base dataset have great differences in their image characteristics.

\subsection{Architecture and training}

Convolutional neural networks (CNN) are a type of feed forward networks where neuron connectivity is biologically inspired by the organization of the animal visual cortex~\cite{Lecun01, Lecun89}. They are variations of multilayer perceptrons with its neurons arranged in 3 dimensions, imposing a local connectivity pattern between near neurons and sharing the weights of the learned filters.
The VGG19 architecture was influenced by AlexNet~\cite{Krizhevsky12}, a CNN presented to the ImageNet competition~\cite{Russakovsky15} previous years. 

\begin{figure}[htb]
	\includegraphics[width=\linewidth]{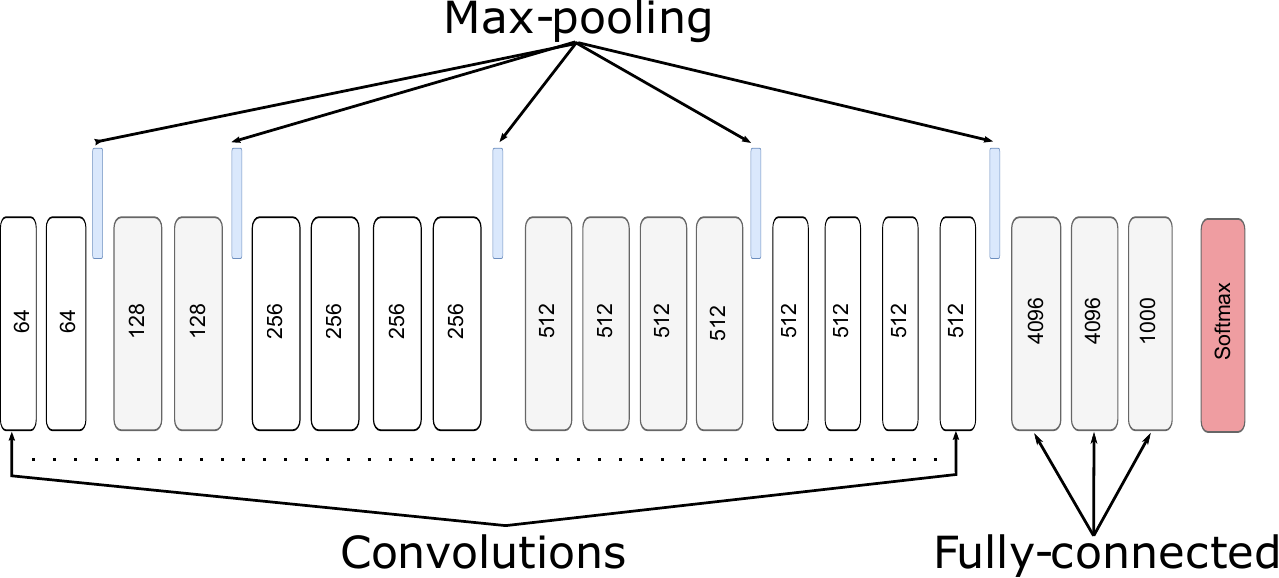}
	\caption{Architecture of the network VGG19. The network has 16 convolutions with ReLUs~\cite{Fürnkranz10} between them and five max-pooling layers. The number of filter maps of the convolutions start at 64 and grow until 512. After the convolutions, there is a linear classifier made-up three fully-connected (fc) layers with dropout~\cite{Srivastava15} between them, the first two have 4096 features while the last one has 1000. The last fc layer is connected to a softmax which maps each value to the probabilities of belonging to each of the 1000 classes of the ImageNet competition.}
	\label{fig:vgg19}
\end{figure}

The VGG19 model has 19 layers with weights (see Figure~\ref{fig:vgg19})), formed by 16 convolutions and 3 fully-connected (fc) layers and its input is an image of size $224\times 224$ and 3 channels with its mean RGB value subtracted.
The convolutional layers have a small kernel size $3\times 3$ with 1 pixel of padding and stride. The network has 5 max-pooling layers with a kernel size of $2\times2$ and stride of 2 pixels. Rectified linear Units (ReLUs)~\cite{Fürnkranz10} are used as the non-linear function. After the convolutional part there is a linear classifier with 3 fully-connected (fc) layers and dropout~\cite{Srivastava15} between them, first two fc layers have 4096 features while the last one has only 1000. The last fc layer is followed by a softmax layer with the same number of outputs which gives the probabilities of the input to belong to each of the 1000 classes of the ImageNet~\cite{Russakovsky15} dataset.

The network was trained using the mini-batch stochastic gradient descent algorithm~\cite{Lecun01}. The batch size was 256 and momentum was set to 0.9. They use the $L_2$ regularization to penalize large weights by a factor of $5\times10^{-4}$. The dropout between the fc layers had a probability of $50\%$ to stop the activations. The learning-rate started at $10^{-2}$ and it was reduced when the loss function stopped to decrease. The training took from 2 to 3 weeks depending on the network. They trained and initialized the weights gradually, starting by random initialization in a network with fewer layers and, once trained, transferring the parameters to a bigger network repeating this process until they reached the network with 19 layers.
In 2011 (before CNNs appeared) a top-5 precision of 75\% was considered a good rate, next year AlexNext~\cite{Krizhevsky12} was the first CNN winning the ImageNet challenge with a top-5 precision of 84\%. The year 2014 the network VGG19 reached 92.7\% top-5 precision obtaining one of the best results in the competition.

\subsection{Evaluation in our dataset}

The network VGG19 has a great performance with natural images; however, neural networks do not always work as expected when the target is not similar to the base dataset. We use the test data of the curated dataset to evaluate the performance of VGG19 obtaining \textbf{26.5\%} and \textbf{47.40\%} of top-1 and top-5 accuracy respectively (see Table~\ref{tab:results-vgg}). The network has more than $40\%$ error in both metrics if we compare it with the precision obtained in natural images. The performance in the noisy dataset is the worst with barely 12\% of top-5 precision due to the labelling problems with some of the images. Its previous training on photographs makes it not capable of recognizing the image characteristics of the new data. The edges, colours and textures differ from the representation that VGG19 has learned with the ImageNet dataset. Also, the degree of abstraction given by the artist to each clip art can make the classification task more complex as shown in Figure~\ref{fig:teaser}.

\begin{table}[htb]
	\centering
	\begin{tabular}{l*{3}{|c}|}
		
		\cline{2-4}
		\multicolumn{1}{l|}{}               & ImageNet & Noisy & Curated \\ \hline
		\multicolumn{1}{|l|}{Prec. top-1}   & 75.30 & 4.80 & 26.50 \\ \hline
		\multicolumn{1}{|l|}{Prec. top-5}   & 92.70 & 12.20 & 47.40 \\ \hline
		\noalign{\vskip 3mm} 
			
	\end{tabular}
	\caption{Results on each test subset of each dataset using the network VGG19 pre-trained on ImageNet. The table shows how the network has a great performance with natural images while its precision drastically decreases with the illustration images of the Noisy and Curated datasets.}
	\label{tab:results-vgg}
\end{table}

	\section{Proposed models}  \label{sc:method}
\label{sc:method} 

As we have shown, the accuracy obtained with the deep network VGG19 in our illustration dataset drops drastically in comparison with natural images. The main reason is that the statistics of the images in our datasets are different to the original images.
One idea to improve performance in our data would be to create a new convolutional network and train it from scratch. However, it is not a good idea for two reasons: first, we lack the amount of data used to train VGG19, and second, we would be losing all the information that the model already learned. 

In order to tackle this problem, we get inspiration by previous work in transfer learning~\cite{Razavian14, Lin11, Donahue13} and evaluate two new models.
In our first model (Section~\ref{sc:method-1}), we extract the high-level features of the CNN and use them as image descriptors to train a Support Vector Machine (SVM). In our second model (Section ~\ref{sc:method-2}), we additionally reset the low-level layers of the VGG19 network and optimize them with our dataset. The high-level layers, those that in theory are meant to capture shapes and objects, are left almost unchanged.

\subsection{Baseline VGG19 + SVM}
\label{sc:method-1}


A Support Vector Machine (SVM) is a supervised algorithm used for classification and regression. The SVM tries to find the optimal hyperplane that categorizes the classes with the maximum margin between samples of different classes. One kind of SVM uses a non-linear kernel to map the data to a higher dimensional space before finding the optimal hyperplane. In our work, we use the non-linear SVM because their effectiveness when features are quite large and their robustness since they maximize the margin between different data samples. Besides, it does not need as much data as a deep network, so it can be trained with our \textit{curated} dataset.

\paragraph*{Training and Evaluation} We split the \textit{curated} dataset as follows: $16\%$ of the data as validation, $64\%$ as training and 20\% as test data. 
For each image, we obtain a feature vector by taking the second fully-connected layer of the network VGG19.
By using three-fold cross-validation we found that the best performance was given by the radial basis function kernel (RBF) that uses the square of the euclidean distance, the slack variable $C=1$ which allows some errors during training -consequently giving the classifier more flexibility and stability-, and $\gamma=0.0001$ that is the weight of each sample during the training process.
The decision function to train the SVM is one-versus-rest (OVR), it trains one classifier for each class finding the optimal hyperplane that places the samples of the class on one side of the optimal hyperplane while the rest on the other side with the maximum margin between the closest samples of different classes. After training, the top-1 and top-5 precision have increased to \textbf{62.04\%} and \textbf{85.64\%} respectively. The image descriptors obtained from the VGG19 are capable of obtaining better results thanks to the powerful non-linear mapping of the RBF function and the SVM classification. Nevertheless, the
network is still not able to recognize the low-level characteristics of the illustration images which tell us that there is still room for improvement if the network is optimized with the illustration image statistics.

\section{Adaptive layer-based optimization (Optimized VGG19)}
\label{sc:method-2}

In light of the results of the first model, we propose another network trained directly with the illustration images. In a straightforward experiment we took the network VGG19 and continued its training by keeping the original weights unchanged. However, after several epochs, we realized that the cross-entropy error was not reducing. We presumed that the low-level image characteristics could not be understood by the learned parameters, thus generating poor predictions and not reducing the cross-entropy error.
Then, we propose a method to refine the parameters by optimizing the network. Zeiler and Fergus~\cite{Zeiler14} tried to understand deep neural networks through visualizations of the activations that each layer yield. They show that networks recognize low-level image features in their first layers while they are capable of perceiving high-level concepts in their last layers. Inspired by that idea, we propose to adaptively optimize the network taking into account the differences of the target (illustration) and base dataset (natural images). 

If we compare a clip art image with a natural image of the same object we can perceive that the high-level concept, the parts of the object, remains the same in both images; however, the low-level features like colours, textures, and edges have changed (see Figure~\ref{fig:teaser}).  

The problem with deep networks is that they need large amounts of data to be trained. For this reason, although we aim to achieve high performance on the \textit{curated} dataset, the adaptive optimization is done with the \textit{noisy} dataset; its huge size favours the learning of the low-level layers. 

During training, we minimize the following cross-entropy error: 

\begin{equation}
\mathcal{L}(S, L) = - \sum_{i \in N} L_{i} \log (S_i)
\label{eq:crossentropy}
\end{equation} 
where $L$ is a vector containing the groundtruth classes, codified using one-hot encoding, and $S$ is a vector with the probabilities of the input to belong to each class. $N$ is the number of images of the dataset.

We tried several model configurations before we found the one that worked for our particular problem. The model configurations that resulted unsuccessful aimed to make the network understand the new image characteristics of the illustration domain. We tried to restart the weights of the lower layers to learn the low-level characteristics of the new domain, to block the parameter updates on the higher layers avoiding to loose the high level object representation already acquired with the previous training, to add new layers to the network and to optimize the training parameters like dropout, learning rate or momentum. 

\subsection{Optimized VGG19: training and evaluation}

\begin{figure}[htb]
	\centering
	\includegraphics[width=1\linewidth]{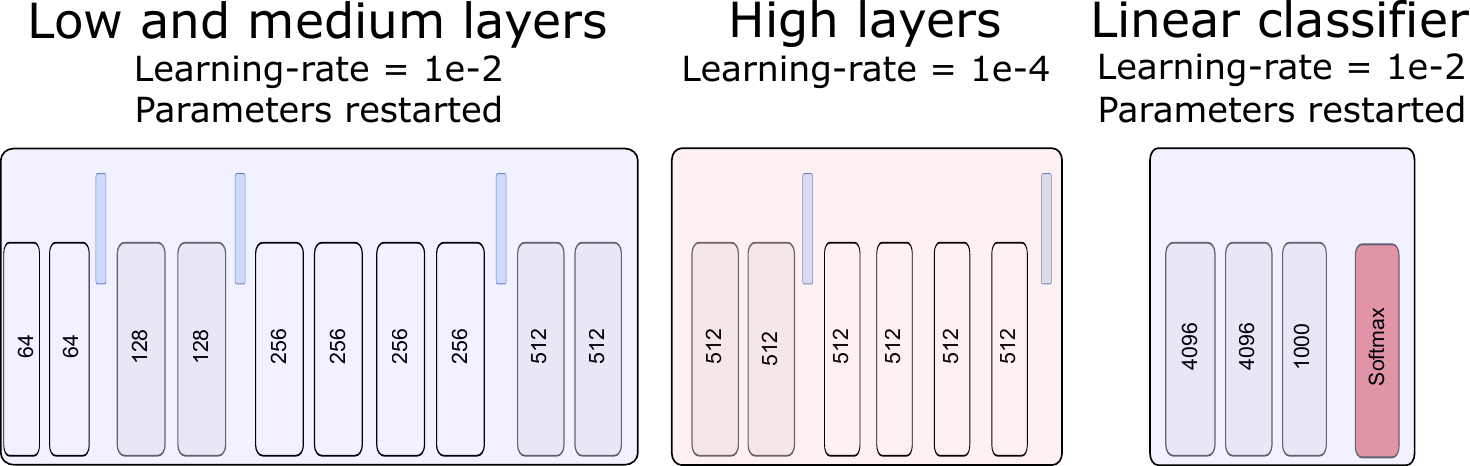}
	\caption{Scheme of the model configuration during adaptive optimization. The lower and medium layers of the network together with the linear classifier have been restarted and their learning rate has been set to a high value allowing the network to learn the new low-level image characteristics. The higher layers of the network remains almost the same, only changing the learning rate to a smaller value not letting the parameters to update substantially.}
	\label{fig:ftScheme}
\end{figure}

The successful model is a combination of the ones we tried previously. We restart the parameters of the low and medium layers ($1^{\text{st}}$ to $10^{\text{th}}$ layer) together with the linear classifier ($17^{\text{th}}$ to the $19^{\text{th}}$ layer) of the VGG19.  The learning rate on these layers is set to $10^{-2}$ allowing the network to learn the new low-level features of the illustration domain and how to classify them forgetting the acquired representation from the ImageNet dataset. On the rest of the layers ($11^{\text{th}}$ to the $16^{\text{th}}$ layer) we set a learning rate of $10^{-4}$ not letting the parameters to be highly modified during the layer-based optimization but with some freedom for them to adapt to the new image domain, a scheme of the model configuration is shown in Figure~\ref{fig:ftScheme}. The optimization converged after 55 epochs and 2 days using a graphic card NVIDIA GTX980Ti. 

Once the network is optimized, we can test its performance by taking the output of the softmax layer as the probability to belong to each class. It achieves a $49\%$ and $70\%$ in top-1 and top-5 precision on the curated dataset. The performance has increased significantly in comparison with the results obtained with the network VGG19 pre-trained on ImageNet and but it stills lower than with the baseline VGG19 + SVM (see Table~\ref{tab:results-comparisson}).

\begin{figure*}[t]
	\includegraphics[width=0.9\linewidth]{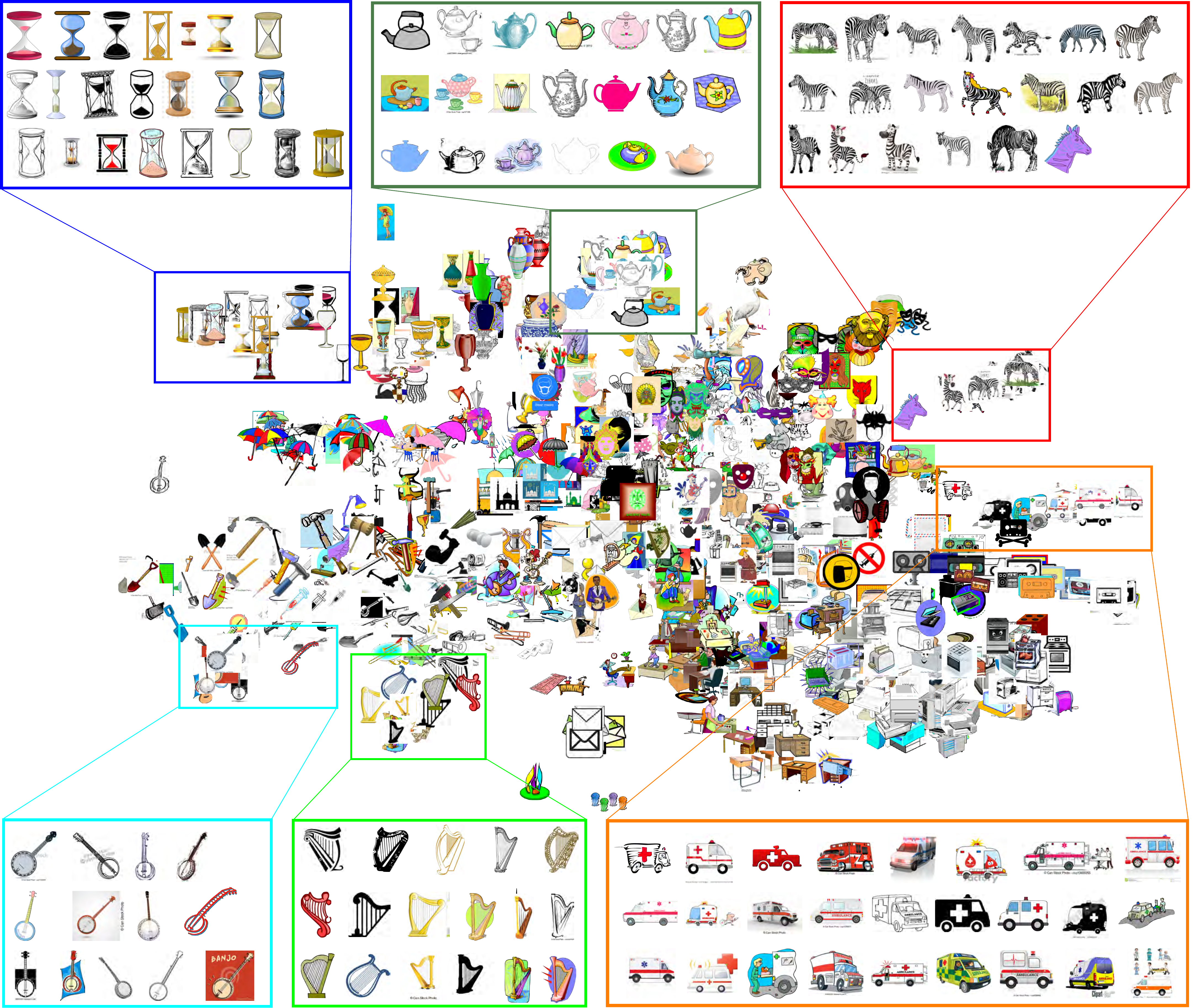}
	\centering
	\caption{t-SNE algorithm on the image descriptors of the optimized network. The boxes show groups of images of the same class that have been grouped together after applying the algorithm. This shows us that the network is able to understand the low-level image characteristics and, consequently, that these image descriptors can be classified with a support vector machine obtaining great results.}
	\label{fig:tsne}
\end{figure*}

We provide a visualization of the learned model by extracting the image-descriptors from the second fully-connected layer of the network. Using the t-SNE algorithm~\cite{Laurens08}, we visualize the image descriptors finding that it has grouped the images of the same class together (Figure~\ref{fig:tsne}).

\subsection{Optimized VGG19 + SVM: training and evaluation}

The optimized VGG19 with the softmax layer as a classifier is not capable of outperforming the results obtained with the baseline VGG19 and the SVM. In Section~\ref{sc:method-1} we have shown how the use of a support vector machine can yield to accurate results; therefore, we train a new SVM with the image descriptors of the curated dataset extracted from the optimized network. We use three fold cross-validation to find the best parameters and the data is split as mentioned before. The best parameters we obtain are the sigmoid kernel, $C=10$, $\gamma=0.0001$ and the decision function one-vs-rest (OVR). The new model obtains a great performance, with \textbf{86.61\%} and \textbf{97.21\%} of top-1 and top-5 global precision, improving the previous model by around 20\% in precision top-1 and 10\% in top-1. A comparison of the precision obtained with the \emph{Optimized VGG19 + SVM} and the \emph{baseline VGG19 + SVM} can be found in Table~\ref{tab:results-comparisson}. The individual accuracies per class can be seen in Table~\ref{tab:results}.

\begin{table}[!h]
	\centering
	
	\begin{tabular}{l*{2}{|c}|}
		\cline{2-3}
		\multicolumn{1}{l|}{}               		& Prec. top-1 & Prec. top-5 \\ \hline
		\multicolumn{1}{|l|}{Baseline VGG19}   		& 26.50 & 47.40 \\ \hline
		\multicolumn{1}{|l|}{Baseline VGG19 + SVM}  & 62.04 & 85.64 \\ \hline
		\multicolumn{1}{|l|}{Optimized VGG19}   	& 49.39 & 70.07 \\ \hline
		\multicolumn{1}{|l|}{Optimized VGG19 + SVM}    		& 86.61 & 97.21 \\ \hline
	\end{tabular}
	\caption{Global top-1 and top-5 precision for each of the proposed models using the curated dataset.}
	\label{tab:results-comparisson}
\end{table}

\begin{figure*}[t]
	\includegraphics[width=\linewidth]{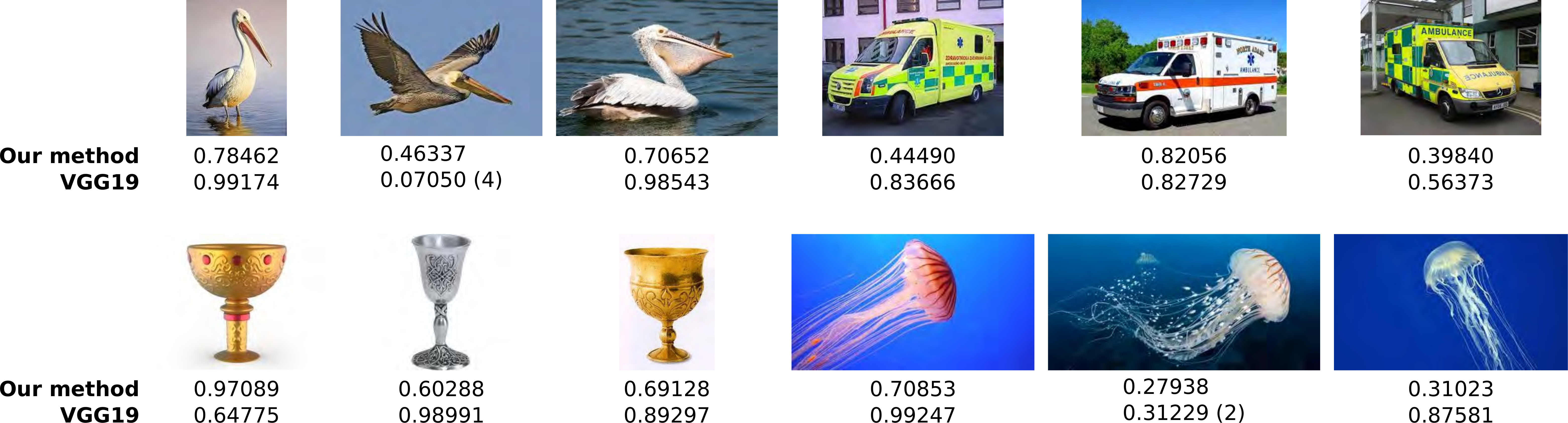}
	\caption{Comparison of the results obtained with the \textbf{optimized model} and the \textbf{baseline VGG19} on randomly sampled natural images. The optimized model returned the correct class as the first prediction in all of them while the baseline VGG19 failed to predict two photographs. It predicted one pelican as the $4^{\text{th}}$ class and a jellyfish as the 2$^{\text{nd}}$, which is stated in the figure by placing the prediction place between brackets next to the probability. Although the optimized network has been able to predict correctly all the images, we can see that some of the photographs (like the last two jellyfish) have been predicted with a relatively low probability. This behaviour is different on the baseline VGG19 where we can see that if the network predicts the image correctly its probability is really high.}
	\label{fig:testPhotos}
\end{figure*}

\paragraph*{Evaluation on natural images} We also analyze how our model performs on natural images. As shown in Figure~\ref{fig:testPhotos} the network is still able to generate good results for photographs.
The adaptive layer optimization has kept the higher layers of the network with slight variations compared to the baseline VGG19, therefore the network is still able to perform acceptably on natural images

\begin{figure}[!h]
	\includegraphics[width=0.8\linewidth]{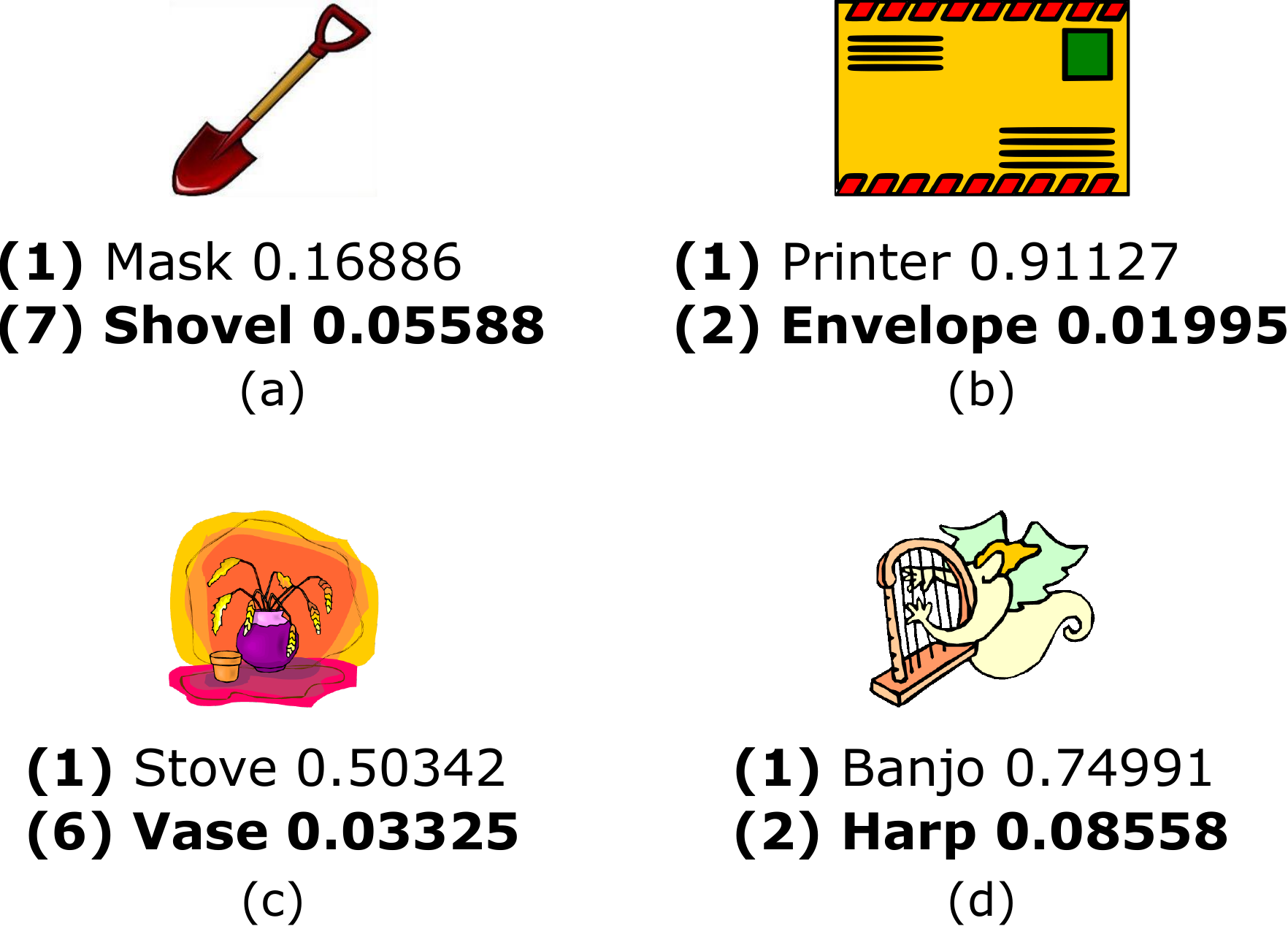}
	\centering
	\caption{Illustrations incorrectly classified by the optimized model from the curated dataset. (a) The model did not clearly know which class is the correct one, predicting it as a mask with low probability. (b) The image of the envelope has a different behaviour, its features are making the model think that it is a printer with high probability. The same is happening with the image (d). (c) We observe that its background could resemble a stove light on which could be confusing the network and making it predict the image as a stove instead of focusing on the vase at the foreground.}
	\label{fig:errorCliparts}
\end{figure}

\paragraph*{Failure cases} In Figure~\ref{fig:errorCliparts} we show a set of clip arts where the proposed optimized model did not work as expected predicting the wrong class. Some of the images where the model does not work contain features that are not powerful to make it know which objects are present; thus returning low probabilities for all the classes. In other cases, the image contains powerful characteristics that the network thinks that belong to another class e.g. the shape or colours could resemble other class. 

\begin{table*}[t]
\centering
\begin{tabular}{l*{8}{|c}|}

\multicolumn{1}{l}{} & \multicolumn{8}{c}{\large\textbf{OPTIMIZED VGG19 + SVM}}\\ [0.7ex]
\cline{2-9}
\multicolumn{1}{l|}{}               & Ambulance & Banjo & Cassette & Desk  & Envelope & Goblet & Hammer & Harp \\ \hline
\multicolumn{1}{|l|}{Prec. top-1}   & 93.10 & 100.0 & 92.30 & 84.62 & 74.07 & 85.19 & 75.0 & 84.62 \\ \hline
\multicolumn{1}{|l|}{Prec. top-5}   & 100.0 & 100.0 & 100.0 & 94.51 & 96.29 & 88.89 & 94.44 & 96.15\\ \hline
\noalign{\vskip 3mm} 

\cline{2-9}
                                   & Hourglass & Jellyfish & Mask & Mosque & Pelican & Printer & Shovel & Stove \\ \hline
\multicolumn{1}{|l|}{Prec. top-1}  & 100.0 & 73.91 & 94.64 & 100.0 & 74.19 & 87.50 & 87.10 & 97.06 \\ \hline
\multicolumn{1}{|l|}{Prec. top-5}  & 100.0 & 100.0 & 100.0 & 100.0 & 96.77 & 97.92 & 96.77 & 100.0 \\ \hline
\noalign{\vskip 3mm} 

\cline{2-9}
								    & Syringe & Teapot  & Toaster & Trombone & Umbrella & Vase & Zebra & \textbf{Global} \\ \hline
\multicolumn{1}{|l|}{Prec. top-1}   & 79.31 & 79.47 & 67.57 & 89.56 & 92.59 & 91.49 & 88.46 & \textbf{86.61} \\ \hline
\multicolumn{1}{|l|}{Prec. top-5}   & 96.55 & 100.0 & 100.0 & 96.55 & 98.15 & 93.62 & 92.31 & \textbf{97.21} \\ \hline

\end{tabular}

\vspace{10mm}

\begin{tabular}{l*{8}{|c}|}

	\multicolumn{1}{l}{} & \multicolumn{8}{c}{\large\textbf{BASELINE VGG19 + SVM}}\\ [0.7ex]
	\cline{2-9}
	\multicolumn{1}{l|}{}               & Ambulance & Banjo & Cassette & Desk  & Envelope & Goblet & Hammer & Harp \\ \hline
	\multicolumn{1}{|l|}{Prec. top-1}   & 44.83 & 80.0 & 46.15 & 40.66 & 74.07 & 77.78 & 64.11 & 88.84 \\ \hline
	\multicolumn{1}{|l|}{Prec. top-5}   & 62.09 & 96.0 & 73.08 & 76.93 & 92.59 & 92.59 & 91.67 & 100.0 \\ \hline
	\noalign{\vskip 3mm} 
	
	\cline{2-9}
	& Hourglass & Jellyfish & Mask & Mosque & Pelican & Printer & Shovel & Stove \\ \hline
	\multicolumn{1}{|l|}{Prec. top-1}  & 70.37 & 65.22 & 51.79 & 83.33 & 32.25 & 66.67 & 64.30 & 73.53 \\ \hline
	\multicolumn{1}{|l|}{Prec. top-5}  & 100.0 & 82.86 & 89.29 & 91.67 & 51.61 & 95.83 & 87.10 & 91.18 \\ \hline
	\noalign{\vskip 3mm} 
	
	\cline{2-9}
	& Syringe & Teapot  & Toaster & Trombone & Umbrella & Vase & Zebra & \textbf{Global} \\ \hline
	\multicolumn{1}{|l|}{Prec. top-1}   & 82.76 & 61.54 & 70.27 & 82.76 & 83.33 & 38.30 & 46.15 & \textbf{62.04} \\ \hline
	\multicolumn{1}{|l|}{Prec. top-5}   & 100.0 & 84.62 & 94.59 & 100.0 & 96.30 & 65.96 & 65.38 & \textbf{85.64} \\ \hline

\end{tabular}

\caption{Comparison between the \textbf{optimized VGG19 + SVM} and the \textbf{baseline VGG19 + SVM} with its precision top-1 and top-5 of each class. In the tables we can observe how the optimized model outperform in almost every class the accuracies of the baseline VGG19 + SVM thanks to the fine-tuning. There are some classes where the predictions in the baseline VGG19 + SVM are slightly better than the ones obtained with the optimized model (class syringe and toaster) but the trade-off still benefits our method.}
\label{tab:results}
\end{table*}

	\section{Conclusions} \label{sc:conclusion}

In this work, we have explored how transferable are the high-level layers of a deep neural network in two different domains which are natural images and illustrations.
We have presented a new illustration dataset with labelled and curated data. 
We have shown that a deep neural network trained for natural images fails when classifying a target dataset with more abstract depiction such as cartoons or clip arts.
We have proposed two models that improve performance by 30-60\% respectively over the original network, and we have shown take our model is still able to work reasonably well on photographs.

There are many interesting avenues of future work. The Curated dataset leaves out most of the available data in the Noisy dataset, using only 23 classes out of 826. This could be improved by using a crowd-sourcing platform to curate our data, as currently, it is a manual process. We would like to perform further experiments to evaluate exhaustively the accuracy of our network in the ImageNet dataset, as our current experiments suggest that the concepts remain in the network. A very interesting problem would be to investigate about abstraction and perception in this kind of networks, for example, to find out if the Gestalt laws are automatically learned by this models, or to predict which are the essential edges or parts that make us recognize an object.

\section{Acknowledgments}
This research has been partially funded by an ERC Consolidator Grant (project CHAMELEON), and the Spanish Ministry of Economy and Competitiveness (projects TIN2016-78753-P and TIN2016-79710-P).
	
	\bibliographystyle{eg-alpha-doi}
	\bibliography{cartoons}

\newcommand{\etalchar}[1]{$^{#1}$}
\begin{thebibliography}{\uppercase{EEVG{\etalchar{*}}15}}

\bibitem[BSCL14]{Babenko14}
\textsc{Babenko A., Slesarev A., Chigorin A., Lempitsky V.}:
\newblock Neural codes for image retrieval, 2014.

\bibitem[CAE{\etalchar{*}}16]{Christodoulidis16}
\textsc{Christodoulidis S., Anthimopoulos M., Ebner L., Christe A., Mougiakakou
  S.~G.}:
\newblock Multi-source transfer learning with convolutional neural networks for
  lung pattern analysis.
\newblock \emph{CoRR abs/1612.02589} (2016).

\bibitem[CPP{\etalchar{*}}17]{Cheplygina17}
\textsc{Cheplygina V., Pe{\~{n}}a I.~P., Pedersen J. J.~H., Lynch D.~A.,
  S{\o}rensen L., de~Bruijne M.}:
\newblock Transfer learning for multi-center classification of chronic
  obstructive pulmonary disease.
\newblock \emph{CoRR abs/1701.05013} (2017).

\bibitem[CZ14]{Crowley14a}
\textsc{Crowley E.~J., Zisserman A.}:
\newblock In search of art.
\newblock In \emph{European Conference on Computer Vision (ECCV)} (2014).

\bibitem[DJV{\etalchar{*}}13]{Donahue13}
\textsc{Donahue J., Jia Y., Vinyals O., Hoffman J., Zhang N., Tzeng E., Darrell
  T.}:
\newblock Decaf: {A} deep convolutional activation feature for generic visual
  recognition.
\newblock \emph{CoRR abs/1310.1531} (2013).

\bibitem[EEVG{\etalchar{*}}15]{Everingham15}
\textsc{Everingham M., Eslami S. M.~A., Van~Gool L., Williams C. K.~I., Winn
  J., Zisserman A.}:
\newblock The pascal visual object classes challenge: A retrospective.
\newblock \emph{International Journal of Computer Vision 111}, 1 (jan 2015),
  98--136.

\bibitem[GAgH14]{Garces14}
\textsc{Garces E., Agarwala A., gutierrez D., Hertzmann A.}:
\newblock A similarity measure for illustration style.
\newblock \emph{ACM Transactions on Graphics (SIGGRAPH 2014) 33}, 4 (2014).

\bibitem[HZRS15]{He15}
\textsc{He K., Zhang X., Ren S., Sun J.}:
\newblock Deep residual learning for image recognition.
\newblock \emph{CoRR abs/1512.03385} (2015).

\bibitem[KNH]{Krizhevsky-cifar}
\textsc{Krizhevsky A., Nair V., Hinton G.}:
\newblock Cifar-10 (canadian institute for advanced research).

\bibitem[KSH12]{Krizhevsky12}
\textsc{Krizhevsky A., Sutskever I., Hinton G.~E.}:
\newblock Imagenet classification with deep convolutional neural networks.
\newblock In \emph{Advances in Neural Information Processing Systems 25},
  Pereira F., Burges C. J.~C., Bottou L., Weinberger K.~Q., (Eds.). Curran
  Associates, Inc., 2012, pp.~1097--1105.

\bibitem[LBBH01]{Lecun01}
\textsc{LeCun Y., Bottou L., Bengio Y., Haffner P.}:
\newblock Gradient-based learning applied to document recognition.
\newblock In \emph{Intelligent Signal Processing} (2001), IEEE Press,
  pp.~306--351.

\bibitem[LBD{\etalchar{*}}89]{Lecun89}
\textsc{LeCun Y., Boser B., Denker J.~S., Henderson D., Howard R.~E., Hubbard
  W., Jackel L.~D.}:
\newblock Backpropagation applied to handwritten zip code recognition.
\newblock \emph{Neural Computation 1}, 4 (Winter 1989), 541--551.

\bibitem[LLZ{\etalchar{*}}11]{Lin11}
\textsc{Lin Y., Lv F., Zhu S., Yang M., Cour T., Yu K.}:
\newblock Large-scale image classification: Fast feature extraction and svm
  training.
\newblock 2011.

\bibitem[MDZF10]{Zeiler10}
\textsc{Matthew D.~Zeiler Dilip~Krishnan G. W.~T., Fergus R.}:
\newblock \emph{Deconvolutional networks}.
\newblock 2010, pp.~2528--2535.

\bibitem[NH10]{Fürnkranz10}
\textsc{Nair V., Hinton G.~E.}:
\newblock Rectified linear units improve restricted boltzmann machines.
\newblock In \emph{Proceedings of the 27th International Conference on Machine
  Learning (ICML-10)} (2010), Fürnkranz J., Joachims T., (Eds.), Omnipress,
  pp.~807--814.

\bibitem[OBLS14]{Oquab14}
\textsc{Oquab M., Bottou L., Laptev I., Sivic J.}:
\newblock Learning and transferring mid-level image representations using
  convolutional neural networks.
\newblock In \emph{Computer Vision and Pattern Recognition (CVPR)} (2014).

\bibitem[RASC14]{Razavian14}
\textsc{Razavian A.~S., Azizpour H., Sullivan J., Carlsson S.}:
\newblock Cnn features off-the-shelf: an astounding baseline for recognition,
  2014.

\bibitem[RDS{\etalchar{*}}15]{Russakovsky15}
\textsc{Russakovsky O., Deng J., Su H., Krause J., Satheesh S., Ma S., Huang
  Z., Karpathy A., Khosla A., Bernstein M., Berg A.~C., Fei-Fei L.}:
\newblock Imagenet large scale visual recognition challenge.
\newblock \emph{International Journal of Computer Vision (IJCV) 115}, 3 (2015),
  211--252.

\bibitem[SEZ{\etalchar{*}}13]{Sermanet13}
\textsc{Sermanet P., Eigen D., Zhang X., Mathieu M., Fergus R., LeCun Y.}:
\newblock Overfeat: Integrated recognition, localization and detection using
  convolutional networks.
\newblock \emph{CoRR abs/1312.6229} (2013).

\bibitem[SHK{\etalchar{*}}14]{Srivastava15}
\textsc{Srivastava N., Hinton G., Krizhevsky A., Sutskever I., Salakhutdinov
  R.}:
\newblock Dropout: A simple way to prevent neural networks from overfitting.
\newblock \emph{Journal of Machine Learning Research 15} (2014), 1929--1958.

\bibitem[SLJ{\etalchar{*}}15]{Szegedy15}
\textsc{Szegedy C., Liu W., Jia Y., Sermanet P., Reed S., Anguelov D., Erhan
  D., Vanhoucke V., Rabinovich A.}:
\newblock Going deeper with convolutions.
\newblock In \emph{Computer Vision and Pattern Recognition (CVPR)} (2015).

\bibitem[SZ14]{Simonyan14}
\textsc{Simonyan K., Zisserman A.}:
\newblock Very deep convolutional networks for large-scale image recognition.
\newblock \emph{CoRR abs/1409.1556} (2014).

\bibitem[Tan13]{Tang13}
\textsc{Tang Y.}:
\newblock Deep learning using support vector machines.
\newblock \emph{CoRR abs/1306.0239} (2013).

\bibitem[vdMH08]{Laurens08}
\textsc{van~der Maaten L., Hinton G.~E.}:
\newblock Visualizing high-dimensional data using t-sne.
\newblock \emph{Journal of Machine Learning Research 9} (2008), 2579--2605.

\bibitem[YCBL14]{Yosinski14}
\textsc{Yosinski J., Clune J., Bengio Y., Lipson H.}:
\newblock How transferable are features in deep neural networks?
\newblock \emph{CoRR abs/1411.1792} (2014).

\bibitem[ZF14]{Zeiler14}
\textsc{Zeiler M.~D., Fergus R.}:
\newblock Visualizing and understanding convolutional networks.
\newblock In \emph{European Conference on Computer Vision (ECCV)} (2014),
  pp.~818--833.

\end{thebibliography}

\end{document}